\documentclass{article}

\usepackage{arxiv}

\usepackage[utf8]{inputenc} % allow utf-8 input
\usepackage[T1]{fontenc}    % use 8-bit T1 fonts
\usepackage{hyperref}       % hyperlinks
\usepackage{url}            % simple URL typesetting
\usepackage{booktabs}       % professional-quality tables
\usepackage{amsfonts}       % blackboard math symbols
\usepackage{nicefrac}       % compact symbols for 1/2, etc.
\usepackage{microtype}      % microtypography

\usepackage{amsmath} %,amssymb,amsfonts}, \usepackage{amsthm}

\usepackage{cleveref}       % smart cross-referencing
\usepackage{lipsum}         % Can be removed after putting your text content
\usepackage{graphicx}
\usepackage{natbib}
\usepackage{doi}

\usepackage{algorithm}
\usepackage{algorithmicx}
\usepackage{algpseudocode}
\usepackage{subcaption} % subfigure
\usepackage{multirow}

\title{Enhancing Multiple Object Tracking Accuracy \\ via Quantum Annealing}

\newif\ifuniqueAffiliation
\uniqueAffiliationtrue

\author{ Yasuyuki Ihara\\
	Innovation Promotion Division \\
	NEC Solution Innovators, Ltd.\\
	1-18-7 Shinkiba, Koto-ku, Tokyo, 136-8627 Japan. \\
	\texttt{ihara\_mxk[ ]nec.com}\thanks{Replace [ ] with @.}
}

\renewcommand{\headeright}{ }
\renewcommand{\undertitle}{ }
\renewcommand{\shorttitle}{Enhancing Multiple Object Tracking Accuracy via Quantum Annealing}

\hypersetup{
pdftitle={Enhancing Multiple Object Tracking Accuracy via Quantum Annealing},
pdfsubject={ },
pdfauthor={Yasuyuki Ihara},
pdfkeywords={Multiple object tracking, Quantum annealing, Reverse annealing, Combinatorial optimization problem, Image recognition},
}

\begin{document}
\maketitle

\begin{abstract}
Multiple object tracking (MOT), a key task in image recognition, presents a persistent challenge in balancing processing speed and tracking accuracy. This study introduces a novel approach that leverages quantum annealing (QA) to expedite computation speed, while enhancing tracking accuracy through the ensembling of object tracking processes. A method to improve the matching integration process is also proposed. By utilizing the sequential nature of MOT, this study further augments the tracking method via reverse annealing (RA). Experimental validation confirms the maintenance of high accuracy with an annealing time of a mere 3 $\mu$s per tracking process. The proposed method holds significant potential for real-time MOT applications, including  traffic flow measurement for urban traffic light control, collision prediction for autonomous robots and vehicles, and  management of products mass-produced in factories.
\end{abstract}


\def\SOTA#1{fan2019lasot}
\def\SOTB#1{zheng2020learning}
\def\MVA#1{apaydin2008structure}
\def\MVB#1{yoshida2022experimental}
\def\MVC#1{shim2017ising}
\def\MOTA#1{zhang2008global}
\def\MOTB#1{li2009learning}
\def\MOTC#1{bewley2016simple}
\def\MOTG#1{brendel2011multiobject}
\def\MOTD#1{wojke2017simple}
\def\MOTE#1{zhang2022bytetrack}
\def\MOTF#1{dai2019video}
\def\MOTG#1{gebregziabher2023multi}
\def\MOTH#1{gudauskas2021multiple}
\def\MOTI#1{ward2021real}
\def\HUNGA#1{kuhn1955hungarian}
\def\HUNGB#1{munkres1957algorithms}
\def\AQCA#1{farhi2001quantum}
\def\AQCB#1{childs2001robustness}
\def\QMOTA#1{zaech2022adiabatic}
\def\QMOTB#1{mccormick2022multiple}
\def\QMOTC#1{mccormick2022implementation}
\def\COISING#1{lucas2014ising}
\def\QAA#1{kadowaki1998quantum}
\def\EMOTB#1{peng2020dense}
\def\EMOTC#1{du2021giaotracker}
\def\EMOTA#1{du2022ensemblemot}
\def\TTS#1{albash2018adiabatic}
\def\DETRACA#1{uadetrac}
\def\DETRACB#1{wen2020ua}
\def\PCHA#1{zauner2011rihamark}
\def\MOTAA#1{bernardin2008evaluating}
\def\IDFA#1{ristani2016performance}
\def\RAA#1{passarelli2020reverse}
\def\RAB#1{venturelli2019reverse}
\def\RAC#1{haba2022travel}
\def\RAD#1{reverse_annealing}
\def\ministate#1{\begin{minipage}[t]{45zw}#1\end{minipage}}

\section{Introduction}\label{Introduction}

Multiple object tracking (MOT) is a key task in image recognition, holding an advantage over single-object tracking~\cite{\SOTA{2}, \SOTB{2}} in scenarios with crowded detected objects and frequent occlusions. Because of its superiority, it has been the subject of active research in recent years~\cite{\MOTA{2},\MOTB{2},\MOTC{2},\MOTD{2},\MOTE{2},\MOTF{2},\MOTG{2},\MOTH{2},\MOTI{2}}. MOT is also being used in a variety of real-world situations, such as counting the number of traveling vehicles to obtain traffic flow information~\cite{\MOTF{2}}, predicting approaches of nearby objects for autonomous robots~\cite{\MOTG{2}}, and analyzing the movements of players in sports 
videos~\cite{\MOTH{2}}. Not only accuracy in object tracking but also latency-free real-time processing are needed in situations where processing of control is required immediately after tracking. Such cases include traffic flow control in big cities, where traffic flow is measured as input to control traffic light; collision avoidance in autonomous robotics and vehicles~\cite{\MOTG{2}}; and quality control in manufacturing~\cite{\MOTI{2}}., where abnormal objects must be detected from large quantities of manufactured items in factories.

In research to date, MOT has been frequently formulated as a multi-dimensional assignment problem, a type of combinatorial optimization problem (COP). The optimal solution is often obtained by the Hungarian method~\cite{\HUNGA{2},\HUNGB{2}}. However, given that the order is $O(n^3)$ (where $n$ represents the number of tracks), real-time processing becomes difficult as the scale of computation increases. While research has been conducted to enhance tracking accuracy by integrating multiple types of MOT, the suppression of computational costs has not yet been achieved.

In recent years, quantum annealing (QA) has emerged as a new method for solving COPs, and has become an active area of research. Unlike simulated annealing, which searches for optimal solutions using thermal fluctuations, QA is an algorithm that uses quantum fluctuations to search for optimal solutions to COPs. It is a heuristic algorithm that utilizes quantum physics phenomena and has drawn attention for its ability to evaluate all solution combinations simultaneously using the quantum superposition effect, and to pass through potential barriers using the tunneling effect. As such, it can find the global optimal solution quickly and accurately without being trapped in local solutions. A technique analogous to QA, called adiabatic quantum computation (AQC), has also been proposed~\cite{\AQCA{2},\AQCB{2}}.  
In AQC, a system starts from the ground state, and adiabatic time evolution is used to find the combinatorial optimization solution. Both QA and AQC use adiabatic time evolution, but AQC is a logical computational model that can also be implemented using quantum gates.

Research on MOT using QA or adiabatic quantum computation has been carried out in recent years~\cite{\QMOTA{2},\QMOTB{2},\QMOTC{2}}. However, these studies have been limited to the examination of the reduction of computational costs and have not reported improvements in accuracy.

This paper proposes a method that not only accelerates object tracking calculations using  QA, but also enhances accuracy by ensemble processing of object tracking by leveraging the characteristics of QA. 
Another method that deploys reverse annealing (RA) to further boost the efficiency of object tracking processing is also proposed.

The organization of this paper is as follows. Section \ref{sec:MOT} provides an overview of conventional  MOT. Section \ref{sec:QACOPT} discusses the general theory of combinatorial optimization using QA. Section \ref{sec:QUBO} formulates the MOT problem in a manner that can be addressed by QA. Section \ref{sec:QEMOT} introduces the first proposed method, Quantum-Ensemble MOT, which enhances accuracy through ensemble processing and accelerates object tracking. The second method, aimed at further optimizing MOT efficiency using RA, is presented in Section \ref{sec:RAMOT}. Both Sections \ref{sec:QEMOT} and \ref{sec:RAMOT} also describe evaluation experiments using the UA-DETRAC dataset, a benchmark for object detection and tracking.
The work is summarized in Section \ref{sec:CONC}.

\section{Multiple Object Tracking (MOT)}\label{sec:MOT}

This section provides an overview of traditional  MOT. The tracking problem is approached through the paradigm of ``Tracking by Detection". In this method, objects are detected in each individual video frame. Detected objects are then temporally associated through a process that operates independently of the detection process, thereby creating distinct object tracks.

\subsection{Formulation of MOT}

\begin{algorithm}[b]
\caption{Pseudocode for MOT}\label{alg:MOT} 
\begin{algorithmic}[1]
\State Initialize tracks 
\For{each frame} 
\State Obtain new detections from observation data 
\State Construct a bipartite graph denoted as $G = (U \cup V, E)$. ($U$: set of current tracks, $V$: set of new detections, $E$: correspondence between tracks and detections). 
\State Compute the cost of edges (e.g., Euclidean distance, similarity of appearance features, etc.). 
\State Determine the optimal matching using a bipartite graph matching algorithm (e.g., Hungarian method). 
\State Assign new detections to current tracks. 
\State Terminate tracks that could not be assigned to new detections.
\State Start new tracks for detections that could not be assigned.
\EndFor  \\
\Return All tracks 
\end{algorithmic}
\end{algorithm}

The MOT problem is often tackled using bipartite graph matching to establish the correspondence of the same object across consecutive video frames. The general pseudocode for this approach to MOT is presented in Algorithm~\ref{alg:MOT}~\cite{\MOTG{2}}.

\subsubsection{DeepSORT}

This subsection discusses DeepSORT~\cite{\MOTD{2}}, a representative and popular method of MOT in recent years.
The SORT (Simple Online and Realtime Tracking) algorithm, a precursor to DeepSORT~\cite{\MOTC{2}}, associates the same object using the Hungarian method. This is based on the positional relationship between the object detection result in the latest frame and the position prediction from the object detection result in past frames by the Kalman filter. Although it is a simple and fast method, there are challenges such as difficulty in tracking when an object is missed mid-process due to phenomena such as occlusions, or when other phenomena interrupt tracking.

\begin{table}[t]
\caption{Experimental Environment}\label{tbl:expenv}
\centering
\begin{tabular}{cl}\toprule
CPU &  Intel Core i7-7700K, 
4 cores 8 threads 4.20 Ghz \\ 
GPU &  NVIDIA GeForce RTZX 2080 Ti 11GB \\ 
Memory & 32.0GB \\ 
OS &  Windows 10 Enterprise (64bit) \\ \bottomrule
\end{tabular}
\end{table}
\begin{figure}[t]
 \centering
\includegraphics[bb=0 0 552 266, width=13cm]{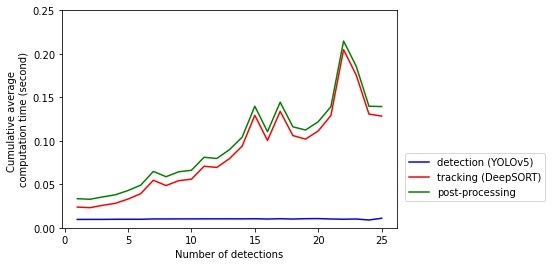}
 \caption{The cumulative processing time for detection by YOLOv5 and tracking by DeepSORT. The right axis represents the number of detected vehicles, and the vertical axis represents the total processing time for detection and tracking (unit: seconds).
}\label{fig:PorpossalMethod01}
\end{figure}

The DeepSORT algorithm, an extension of the SORT algorithm, was developed to address this problem. It first performs matching using appearance features, and for detected objects that did not match, it carries out matching similar to SORT. However, as shown in 
Figure~\ref{fig:PorpossalMethod01}, experiments conducted with DeepSORT in the experimental environment depicted in 
Table~\ref{tbl:expenv} revealed a challenge: the computation time increased as the number of detected objects grew. For instance, when the number of detected objects exceeds 14, achieving detection and tracking at 10fps becomes difficult.

\section{Combinatorial Optimization using Quantum Annealing}\label{sec:QACOPT}

\subsection{Combinatorial Optimization Problems (COPs)}

COPs involve the efficient search for the optimal combination from various possible patterns, using an evaluation function that assesses the quality of a combination. As the size of the problem increases, the total number of combinations also grows exponentially, leading to a phenomenon known as combinatorial explosion. This makes the problem increasingly difficult to solve. Examples of COPs that can be solved in polynomial computation time include the minimum spanning tree problem and the shortest path problem for graphs. Conversely, there are also COPs that require exponential computation time to solve, such as the traveling salesman problem.

Many COPs can be formulated as minimization problems using a cost function $E$ in the form of an Ising model, which is defined with spin variables ($\sigma=\pm 1$)~\cite{\COISING{2}}.

\def\bm{\boldsymbol}

\begin{equation}
\begin{split}
 {\bm \sigma}^{*} &= \underset{\bm \sigma}{\operatorname{argmin}}\ E({\bm \sigma}),  \\
 E({\bm \sigma}) &:= - \sum_{i\neq j} J_{ij} \sigma_i \sigma_j - \sum_i h_i \sigma_i . \label{engising}
\end{split}
\end{equation}
Here, ${\bm \sigma}=(\sigma_1, \cdots, \sigma_N)^\top $ represents a vector of spins, $J_{i,j}$ denotes the interaction variables between spins,
and $h_i$ signifies the strength of an external magnetic field. 
The formulation of combinatorial optimization problems using an Ising model is equivalent to a quadratic binary integer programming problem. This equivalence is achieved through the use of a Quadratic Unconstrained Binary Optimization (QUBO) formulation $E'$ with binary variables $q_i$, which can take on values of $1$ or $0$. The transformation of variables is given by $q_i = \frac{\sigma_i + 1}{2}$, where $\sigma_i$ are the spin variables in the Ising model.
\begin{equation}
\begin{split}
 {\bm q}^{*} &= \underset{\bm q}{\operatorname{argmin}}\ E'({\bm q}),   \\
 E'({\bm q}) &:= - \sum_{i\neq j} a_{ij} q_i q_j - \sum_i b_i q_i .
\end{split}
\end{equation}

\subsection{Basic Principles of QA}\label{subsec:forward-algorithm}

Simulated annealing (SA), which utilizes thermal fluctuations, is recognized as a traditional method for solving COPs. In contrast, 
QA is an algorithm that leverages quantum fluctuations to identify the optimal solution of a COP as the minimum energy solution~\cite{\QAA{2}}. It is capable of traversing potential barriers during the search for the minimum energy solution, thereby enabling a search for the globally optimal solution with the lowest energy without being trapped in local optimum.

In this context, the $x$, $y$, and $z$ components of the Pauli spin matrix $\sigma$, and the $2\times 2$ identity matrix $I$ are defined as follows:
\begin{equation}
\begin{split}
\sigma^{x} & := \begin{pmatrix} 0 & 1 \\ 1 & 0 \end{pmatrix},  \
\sigma^{y} := \begin{pmatrix} 0 & -i \\ i & 0 \end{pmatrix},  \\
\sigma^{z} & := \begin{pmatrix} 1 & 0 \\ 0 & -1 \end{pmatrix}, \
I  := \begin{pmatrix} 1 & 0 \\ 0 & 1 \end{pmatrix}.
\end{split}
\end{equation}
The $n$-fold tensor product of the identity matrix $I$ is denoted as $I^{\otimes n}$, and the $N$-fold tensor product of the Pauli spin matrix $\sigma$ and the identity matrix $I$ is expressed as:
\begin{equation}
\sigma_i^* = I^{\otimes (i-1)} \otimes \sigma^* \otimes  I^{\otimes (N-i)}.  \ \  (*=x,y,z)
\end{equation}

Under the above setting, the Hamiltonian of the Ising model, denoted as
\begin{equation}
H_1 = - \sum_{i\neq j} J_{ij} {\sigma_i}^z {\sigma_j}^z - \sum_i h_i {\sigma_i}^z \label{engising_ham}
\end{equation}
is introduced in relation to the cost function (equation \eqref{engising}) of the COP to be solved. The tensor product $\sigma_i^z$ in equation \eqref{engising_ham} corresponds to the spin variable $\sigma_i$ in equation \eqref{engising}.

To initiate the search for the solution to the COP from a quantum superposition state, the effect of quantum fluctuations due to a transverse magnetic field is utilized to realize a quantum superposition state. The Hamiltonian of the transverse magnetic field is expressed as follows:
\begin{equation}
H_0 = \alpha \sum_i {\sigma_i}^x .
\end{equation}

Subsequently, the Hamiltonian, which integrates $H_0$ and $H_1$, is introduced as:
\begin{equation}
H(t) := \left(1 - \frac{t}{T}\right) H_0 + \frac{t}{T} H_1,  \label{engeq:hamiltonian_forward}
\end{equation}
This Hamiltonian evolves over time from $H_0$ at time $t=0$ to $H_1$ at time $t=T$. 
At the time $t=0$ when the search for the optimal solution begins, a quantum superposition state is realized.
If the time evolution is sufficiently slow, according to the adiabatic theorem of quantum mechanics, in an ideal system free from external noise, the state is always maintained in the ground state. Therefore, it is believed that the ground state of $H_1$, which is desired to be solved, can be obtained with a high probability at time $t=T$~\cite{\QAA{2}}.

\clearpage

\section{Formulation of MOT via QUBO Representation}\label{sec:QUBO}

In this section, the MOT problem described in Section~\ref{sec:MOT} is formulated as the problem of minimum weight maximal matching in a bipartite graph. In existing research on MOT, methods using maximum matching, maximal matching, and stable matching have been proposed. 
However, this paper presents a methodology that employs maximal matching. This approach was chosen for two primary reasons:
\begin{enumerate}
\item Maximal matching can be expressed in the QUBO (Quadratic Unconstrained Binary Optimization) formula mentioned later.
\item In the situation of MOT, the degree of each node in the bipartite graph is often uneven, and compared to the maximum matching prioritizing maximizing the number of matchings, maximal matching tends to yield more stable results.
\end{enumerate}

\begin{figure}[h]
 \centering
 \includegraphics[bb=0 0 650 366, width=7cm]{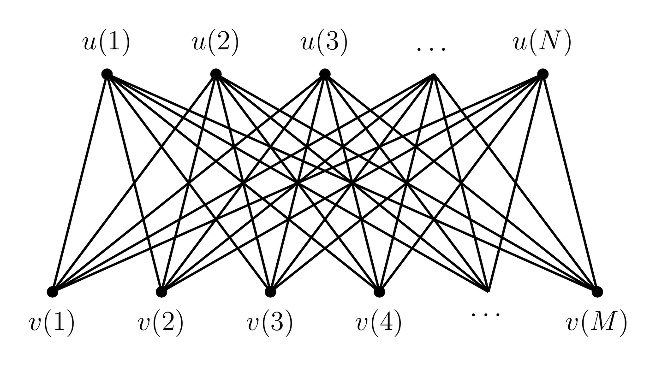}
 \caption{Illustration of matching}\label{fig:matching}
\end{figure}

Let $U=\{ u(i) \}_{i=1}^N$ denote the set of tracks at the detection frame at time $t-1$, and $V=\{v(j) \}_{j=1}^M$ denote the set of detected objects at the detection frame at time $t$. The time $t$ is counted for each detection frame and is assumed to take an integer value. At this time, a bipartite graph $G=(U \cup V, E)$ is introduced, where edges are only placed between $u\in U$ and $v\in V$. Here, $U \cap V = \emptyset$, and $E$ is the set of edges $E = \{ (u(i), v(j) ) \mid u(i) \in  U, \ v(j) \in V  \}$. An illustration of the bipartite graph $G$ is shown in 
Figure ~\ref{fig:matching}.

In MOT problems, when the correspondence of the same object between $U,V$ is represented by $M (\subseteq E)$, different $e_1, e_2\in M (e_1\neq e_2)$ do not share the same node. Therefore, MOT problems can be considered as a type of matching problem in bipartite graph $G$.

To quantitatively evaluate the quality of matching, a function $w$ on $E$ is introduced. The similarity between objects $u(i), v(j)$ in adjacent detection frames is measured by some indicator, such as the distance between objects or the similarity of image features. This indicator is denoted by $w(u(i), v(j))$, which takes values in $[0, 1]$, where a larger value implies a higher similarity.

The MOT problem can now be considered as a maximum-weight maximal matching problem that searches for a matching $M$:
\begin{equation}
M^{*}  = \underset{M}{\operatorname{argmax}} \sum_{m\in M}w(m). \label{engeq:matching}
\end{equation}

Whether or not $(u,v)\in E$ is included in matching $M$ can be represented by
\begin{equation}
x_{u,v} = 
\begin{cases}
1 & (u,v) \in M \\
0 & (u,v) \not\in M
\end{cases}
\end{equation} 
in terms of the binary variable $x_{u,v}$ on $E$. Using the cost function $F_w$ on $E$, 
the optimization problem~\eqref{engeq:matching} can then be rewritten as an optimization problem
\begin{align}\label{engeq:optimization}
{\bm x}^{*} = & \underset{{\bm x} = \{x_{u,v} \mid (u,v) \in E \}}{\operatorname{argmin}}\  F_w ({\bm x}) \nonumber \\
F_w ({\bm x}) := & - \sum_{(u,v)\in E} w(u,v) x_{u,v}
\end{align}  
under constraints
\begin{equation}
\sum_{v\in V}  x_{u,v} \leq 1  \ \ (\forall u\in U), \ \ \
\sum_{u\in U}  x_{u,v} \leq 1  \ \ (\forall v\in V), \label{engeq;conditionsAB}
\end{equation}
which represent matching constraints (matching edges never share the same node).

Here, the variable pair ${\bm x} = \{ x_{u, v} \mid (u, v) \in E \}$ that satisfies the constraint conditions~\eqref{engeq;conditionsAB}
can be considered as one of the sets of variables that minimize the following formulas: 
\begin{equation}
F_U ( {\bm x} ) :=  \frac{ 1 }{ 2 } \sum_{u \in U} \left(  \sum_{v\in V} x_{u,v} - 0.5 \right)^2,  \ \ \
F_V ( {\bm x} ) :=  \frac{ 1 }{ 2 } \sum_{v \in V} \left(  \sum_{u\in U} x_{u,v} - 0.5 \right)^2. \label{engeq:conditionsBB}
\end{equation}
Since the binary variable $x$ always satisfies $x^2=x$, ignoring constant terms allows equations~\eqref{engeq:conditionsBB} to be simplified as the following quadratic homogeneous polynomial:
\begin{equation}
F_U ( {\bm x} ) :=  \sum_{u \in U} \left( \sum_{ i < j} x_{u,v(i) }  x_{u, v(j)} \right), \ \ \ 
F_V ( {\bm x} ) :=  \sum_{v \in V} \left( \sum_{ i < j} x_{u(i),v }  x_{u(j), v} \right).  \label{engeq:conditionsCB}
\end{equation}

Using Lagrange's method of undetermined multipliers, the optimization problem \eqref{engeq:optimization} under the constraints~\eqref{engeq;conditionsAB} can be summarized as 
\begin{align}\label{engeq:QUBO}
{\bm x}^{*} = & \underset{{\bm x} \in \{x_{u,v} \mid (u,v) \in E \}}{\operatorname{argmin}}\  F({\bm x}) \nonumber \\
F ({\bm x}) := & F_w ({\bm x}) + \lambda F_U ({\bm x}) + \lambda F_V ({\bm x})
\end{align}
using constraint expressions~\eqref{engeq:conditionsCB}. Here, $\lambda (>0)$ is a hyperparameter that adjusts the strength of the constraint equation. 
In the subsequent sections, the parameter $\lambda$ is consistently set to $0.7$.
Since $F({\bm x})$ is a QUBO formulation, it can be implemented on a D-Wave machine to solve combinatorial optimization problems.

\clearpage

\section{Proposed Method 1: Quantum-Ensemble MOT to Improve Accuracy}\label{sec:QEMOT}

\subsection{Integration of Multiple Types of Detectors}

To improve the accuracy of MOT, several methods have been developed to integrate the outputs of multiple types of trackers. The Layer-wise Aggregation Discriminative Model (LADM)~\cite{\EMOTB{2}} uses a weighted average of predictors in the detection process, while the GIAOTracker~\cite{\EMOTC{2}} fuses multiple trackers using the score of TrackNMS,
 which is inspired by SoftNMS. Moreover, EnsembleMOT~\cite{\EMOTA{2}}, which is inspired by ensemble learning in machine learning but does not require a model or training, has also been proposed.
However, these developments fail to account for the escalation in computational cost that accompanies the increase in both the number of detected objects and the diversity of trackers.

\subsection{Multiplication of Matching}\label{subsec:multiplematch}

\begin{figure}[b]
 \centering
 \includegraphics[bb=0 0 397 305, width=9.5cm]{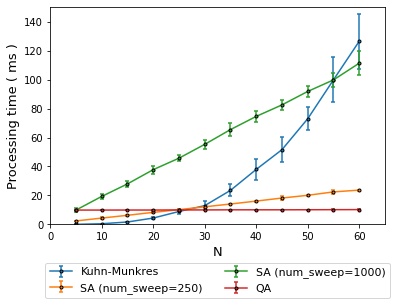}
 \caption{ Number of detected objects and processing time for matching}\label{fig:QPUaccess}
\end{figure}

This paper proposes a method that can reduce the computational cost of bipartite graph matching for object tracking, even when the number of detected objects and the types of trackers increase. The core idea is to use quantum annealing to speed up the search for the optimal matching, as shown in Figure~\ref{fig:QPUaccess}. 
Quantum annealing is utilized to suppress the computation time even when the number of nodes in the bipartite graph increases.

Section~\ref{sec:QUBO} treats the MOT problem as a maximal matching problem in the bipartite graph $G=(U \cup V, E)$ using a single tracker. By employing $F_w$, $F_U$, $F_V$ in Equations \eqref{engeq:optimization} and \eqref{engeq:conditionsBB}, it formulates an optimization problem with the QUBO formula $F$ in Equation~\eqref{engeq:QUBO}, which is solvable by quantum annealing.

\begin{figure}[t]
\centering
 \includegraphics[bb=0 0 1000 400, width=12.0cm]{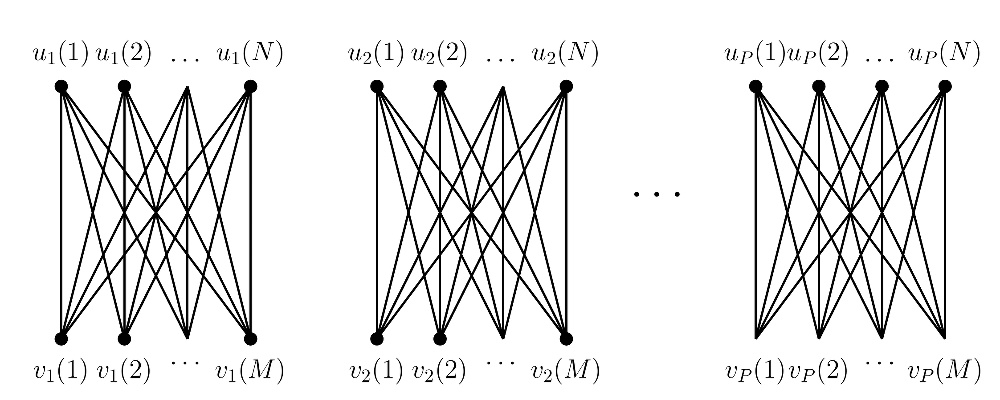}
 \caption{ Illustration of multiplexing matching}\label{fig:multiplematching}
\end{figure}

Considering the scenario where there are $P$ types of trackers $\{ {\cal T}_p \}_{p=1}^P$, each of which can be formulated as maximal matchings in a bipartite graph, an extended bipartite graph $\widehat{G} = \left(\bigcup_{p=1}^P U_p  \cup \bigcup_{p=1}^P V_p, \bigcup_{p=1}^P E_p \right)$ is constructed. This graph duplicates the set of tracks $U=\{ u(i) \}_{i=1}^N$ in the detection frame at time $t-1$ and the set of detected objects $V=\{v(j) \}_{j=1}^M $ in the detection frame at time $t$ into $P$ copies. Here, $E_p$ is a set of edges drawn only between $U_p, V_p$, and the weight $w_p$ of $E_p$ is determined according to the type of tracker ${\cal T}_p$.
Multiplication of Matching is illustrated in Figure ~\ref{fig:multiplematching}.

The problem of $P$ types of tracking is considered, which is combined into a single QUBO equation and is to be solved using quantum annealing:
\begin{align}\label{engeq:integQUBO}
F_p ({\bm x}^p ) := & F_{w_p} ({\bm x}^p ) + \lambda F_{U_p} ({\bm x}^p ) + \lambda F_{V_p} ({\bm x}^p ), \nonumber \\
({\bm x}^1, {\bm x}^2, \cdots, {\bm x}^P )^{*} = & 
\underset{{\bm x}_p \in \{x_{u,v}^p \mid (u,v) \in E_p \}}{\operatorname{argmin}}\  \sum_{p=1}^P F_p ({\bm x}^p ).
\end{align} 
For the detected object $v(j)\in V=\{v(j) \}_{j=1}^M$ at time $t$, the optimal solution  $({\bm x}^1, {\bm x}^2, \cdots, {\bm x}^P )^{*}$ of optimization problem~ \eqref{engeq:integQUBO}  yields a set of track candidates 
\begin{equation}
\{ u_p (i) \mid x_{u_p (i),v_p (j)}^p = 1, 1\leq i \leq N, 1\leq p \leq P  \}
\end{equation} 
related to $v(j)=v_p(j)$. Using the integration method of multiplexed matching described in the next 
subsection~\ref{subsec:integ_matching}, the track candidates are considered for aggregation into a single $i^*$. Then, the track $u(i^*)$ at time $t-1$ is associated with the detected object $v(j)$ at time $t$.

\subsection{Integration of Multiplexed Matching}\label{subsec:integ_matching}

\begin{figure}[h]
  \begin{minipage}[b]{0.45\hsize}
    \centering
    \includegraphics[bb=0 0 650 596, width=4.cm]{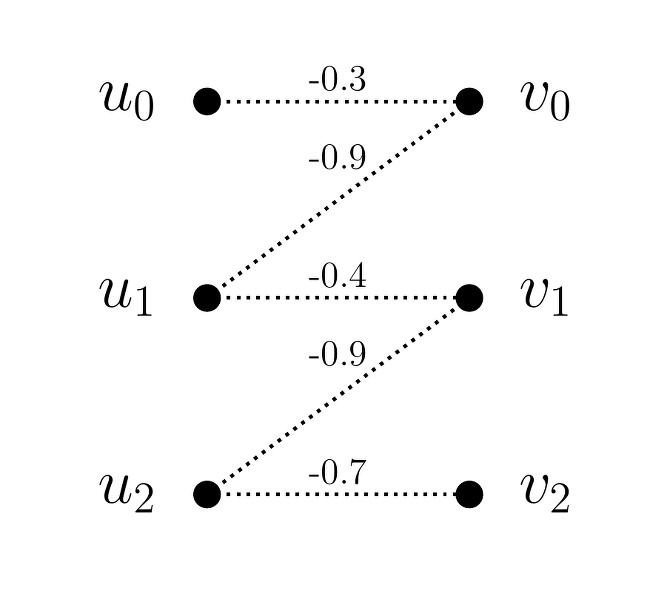}
    %\subcaption{キャプション1}\label{ラベル1}
  \end{minipage}
  \begin{minipage}[b]{0.45\hsize}
    \centering
    \includegraphics[bb=0 0 650 484, width=6.3cm]{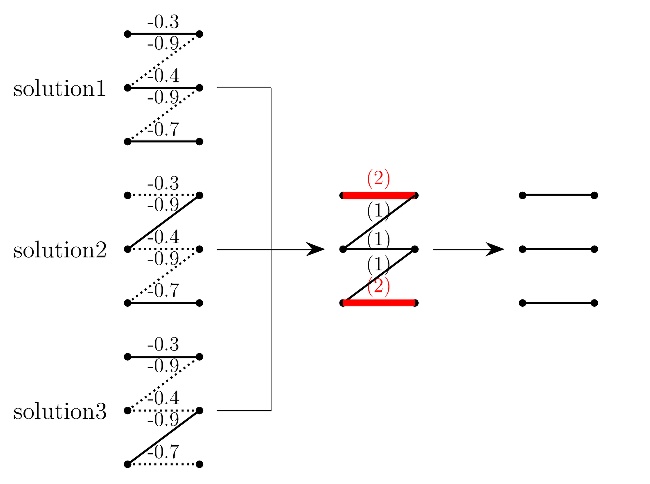}
    %\subcaption{キャプション2}\label{ラベル2}
  \end{minipage}
  \caption{This is an illustration of the integration of multiple matchings by majority voting. For a bipartite graph on the left, three types of matching solutions (solution1-3) are obtained. The voting results are shown in the central graph. The numbers in parentheses on the edges of the central graph represent the number of votes, and the red lines indicate the edges with the most votes. The final solution depicted on the right is a solution of maximum matching, but not of maximal matching.}\label{fig:integration_MV}
  \begin{minipage}[b]{0.45\hsize}
    \centering
    \includegraphics[bb=0 0 389 292, width=6.2cm]{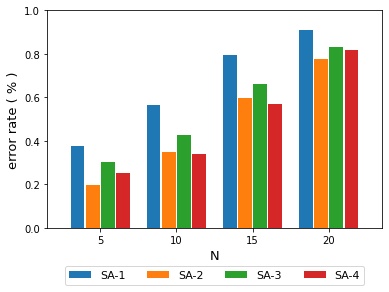}
    \subcaption{SA (sweep=250)}%\label{ラベル1}
  \end{minipage}
  \begin{minipage}[b]{0.45\hsize}
    \centering
    \includegraphics[bb=0 0 389 292, width=6.2cm]{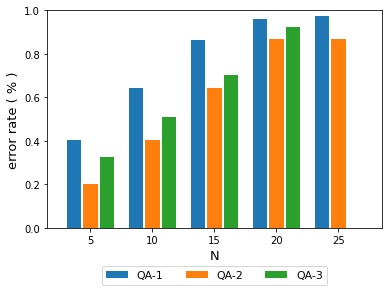}
    \subcaption{QA}%\label{ラベル2}
  \end{minipage}
\caption{Comparison of error rates by majority voting method. The abbreviations SA and QA denote Simulated Annealing and Quantum Annealing, respectively. Furthermore, SA-$k$ and QA-$k$ each represent the integration of solutions for $k$-multiplexed matching.
In cases of triple or higher order matching, satisfactory results have not been achieved in comparison to double matching.
}\label{fig:Error_Rate_MV}
\end{figure}

In conventional research, a method has been adopted to improve the accuracy of solutions by integrating multiple patterns of solutions to combinatorial optimization problems through a majority voting method~\cite{\MVA{2},\MVB{2},\MVC{2}}. In the case of maximal matching problems, the integration by majority voting 
can be showed as in Figure  \ref{fig:integration_MV}.

However, in the case of integration by majority voting, there is an effect of integration if it is a double matching, but no effect of accuracy improvement is observed in the case of triple or more matchings. Ten patterns per $N$ were artificially generated for
$N\times N$ bipartite graph where all nodes have a degree of 4, and the results of the effect of integration of multiple matchings by majority voting were demonstrated by the error rate of the solution in Figure \ref{fig:Error_Rate_MV}. Trial computations were conducted 250 times per graph.

Note that the weights of the edges were randomly assigned in the range of -1.0 to 0.0. The reason for fixing the degree of nodes to 4 is because it is assumed that the average number of neighboring vehicles in the vehicle tracking problem is around 4.

This paper proposes a method for cyclically integrating solutions as follows.
The integration of tracking results by multiple types of tracking methods, as mentioned in the previous subsection, allows for the types of matching problems to be heterogeneous (the same nodes and edges are used for each , but the weights can be different).

For the same bipartite graph $G=(V,E)$ ($V$: set of nodes, $E$: set of edges), weights $W_i$ are assigned to $E$ and denoted as 
$G_i = (V,E,W_i)$. The maximal matching solution of $G_i$ is denoted as $M_i =(V_i, E_i)$ ($V_i$, $E_i$ are subsets of $V$, $E$ respectively).

\begin{algorithm}[h]
\caption{Algorithm for proposed integration method}\label{alg:Integration_Matching} 
\begin{algorithmic}[1]
\State Initialize $K$ with $M_1$.
\For{each iteration $i$ from 1 to $P$}
    \State Integrate $K=(V_K, E_K)$ and $M_j\ (j:=i+1 \mod P)$ to obtain a subgraph $H_i= (V_K \cup V_j, E_K \cup E_j)$ of $G$.
    \If{a path containing a node of degree 2 is found within $H_i$}
        \State Locate an alternating path that minimizes the sum of weights, analogous to finding the solution for maximum matching through augmenting path search.
        \State Establish this as the new integrated solution $K$.
    \EndIf
\EndFor
\end{algorithmic}
\end{algorithm}

\begin{figure}[t]
\centering
\includegraphics[bb=0 0 650 312, width=10.0cm]{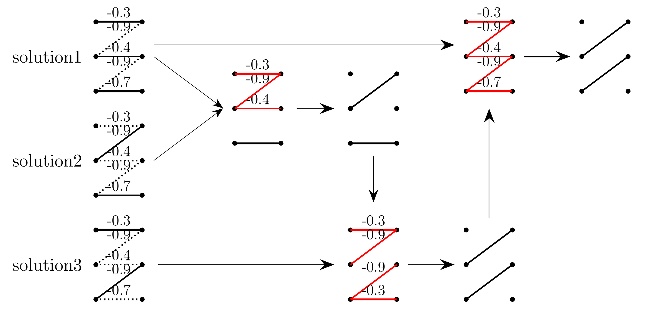}
\caption{This is an illustration of the integration of multiple matchings by the proposed method. Similar to Figure \ref{fig:integration_MV}, when three types of matching solutions (solution1-3) are obtained, first, solution1-2 are integrated. As a result, paths containing nodes of degree 2 that appear are indicated by red lines. Among the red lines, the alternating path whose sum of weights is minimal is retained as the integrated solution. Ultimately, the correct solution of maximal matching is obtained.}\label{fig:integration_Proposed}
\end{figure}

Then, as described in Algorithm~\ref{alg:Integration_Matching} 
and Figure ~\ref{fig:integration_Proposed}, the set of maximal matching solutions $\{ M_i \}_{i=1}^P$ is integrated cyclically.

\begin{figure}[t]
  \begin{minipage}[b]{0.45\hsize}
    \centering
    \includegraphics[bb=0 0 389 292, width=6.25cm]{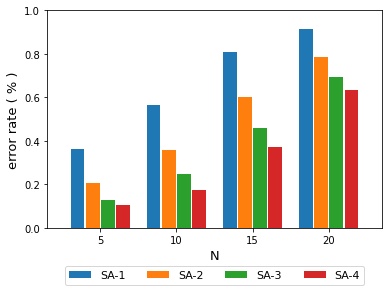}
    \subcaption{SA (sweep=250)}%\label{label1}
  \end{minipage}
  \begin{minipage}[b]{0.45\hsize}
    \centering
    \includegraphics[bb=0 0 389 292, width=6.25cm]{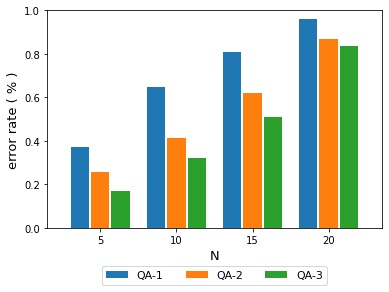}
    \subcaption{QA}%\label{label2}
  \end{minipage}
  \caption{Comparison of error rates by proposed method. Contrary to the case presented in Figure  \ref{fig:Error_Rate_MV}, 
the effect of integration becomes more pronounced as the multiplicity of the matching increases.}\label{fig:Error_Rate_Proposed}
  \begin{minipage}[b]{0.45\hsize}
    \centering
    \includegraphics[bb=0 0 402 288, width=6.25cm]{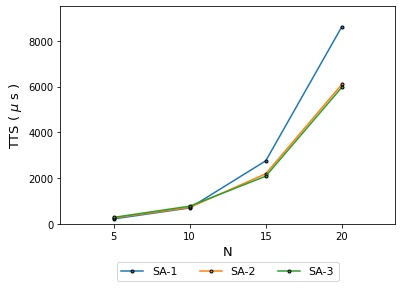}
    \subcaption{SA (sweep=250)}%\label{label1}
  \end{minipage}
  \begin{minipage}[b]{0.45\hsize}
    \centering
    \includegraphics[bb=0 0 402 288, width=6.25cm]{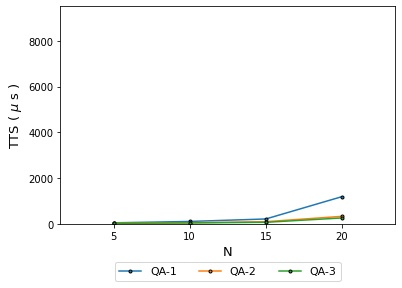}
    \subcaption{QA}%\label{label2}
  \end{minipage}
\caption{Comparison of TTS by proposed method}\label{fig:TTS_Proposed}
\end{figure}

The results of the proposed method's effect on the integration of multiple matchings, verified by the error rate of the solution using the same sample as the experimental results in Figure ~\ref{fig:Error_Rate_MV}, are shown in Figure ~\ref{fig:Error_Rate_Proposed}. In the case of double matching, the accuracy is the same and settings as that of the previously mentioned majority vote integration. However, as the multiplicity of matching increases, the effect of integration becomes apparent.
Furthermore, the effect of matching integration, measured by TTS, is shown in Figure ~\ref{fig:TTS_Proposed}.

\paragraph{TTS}

TTS (Time-to-solution)~\cite{\TTS{2}} is utilized as a metric to measure the efficiency of the simulated or quantum annealing algorithm in finding the optimal solution. TTS signifies the average computation time required to obtain the true optimal solution, given that the annealing time $T$ per 
trial and the target probability $P_R$ of obtaining the true optimal solution at least once are predetermined. When the probability $P_S (t)$ of a successful annealing at time $T$ is considered, TTS is defined as follows:
\begin{equation}
{\rm TTS} := 
\begin{cases}
T \cdot \frac{\log{(1-P_R)}}{\log{(1-P_S (t))}}  & {\rm as}\ P_S (t) < P_R     \\
T & {\rm as}\ P_S (t) \geq P_R
\end{cases}
\end{equation}

Even when verified with TTS, as shown in Figure \ref{fig:TTS_Proposed}, it was found that the effect of multiple matching is significantly demonstrated by the method using QA, which has a less increase in QPU access time.

\subsection{Evaluation Experiments}\label{subsec:evalexp}

This subsection performs a quantitative evaluation of MOT using both conventional baseline methods and the proposed method, thereby demonstrating the superiority of the proposed method.

\subsubsection{Experimental Data}

\begin{figure}[t]
  \begin{minipage}[b]{0.45\hsize}
    \centering
    \includegraphics[bb=0 0 500 281, width=6.2cm]{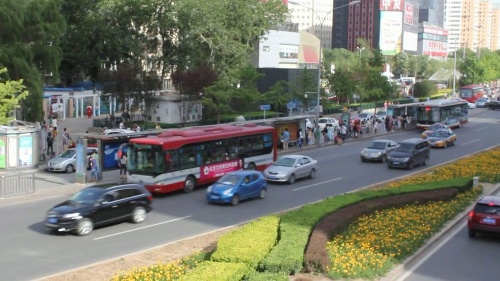}
    \subcaption{MVI 39271}%\label{label1}
  \end{minipage}
  \begin{minipage}[b]{0.45\hsize}
    \centering
    \includegraphics[bb=0 0 500 281, width=6.2cm]{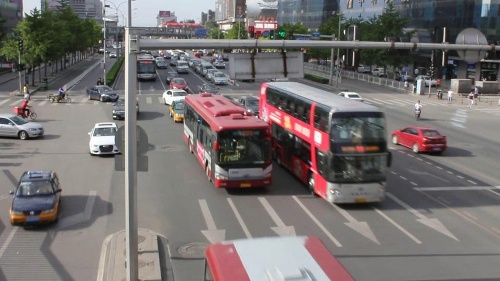}
    \subcaption{MVI 39401}%\label{label2}
  \end{minipage}
  \caption{Overview of UA-DETRAC dataset}\label{fig:UADTRAC}
\end{figure}

The following data is utilized for the experiment:
\begin{itemize}
\item MVI 39271 (1min 18sec, 20fps) and MVI 39401 (1min 9sec, 20fps) from the UA-DETRAC dataset~\cite{\DETRACA{2}, \DETRACB{2}}, which is a benchmark for traffic MOT with pre-annotated vehicle locations and IDs (Figure ~\ref{fig:UADTRAC} ).
\end{itemize}

\subsubsection{Experimental Settings}\label{subsubsec:setting}

Experiments were conducted using the D-Wave Advantage2 prototype1.1  for quantum annealing in the environment outlined in Table~\ref{tbl:expenv}.
For vehicle detection, YOLOv5 was employed across all experimental patterns. As depicted in Figure ~\ref{fig:PorpossalMethod01}, 
the detection process was carried out at approximately 10 frames per second.
For the tracking methods, a comparative evaluation was conducted on the following methods.
\begin{itemize}
\item Baseline Method: DeepSORT, which matches by similarity of appearance features of vehicles detected in the previous frame and matches in bounding box with location prediction.
\item \begin{minipage}[t]{15.0cm}
Proposed Method: As presented in subsection~\ref{subsec:multiplematch}, a method is proposed to solve multiplex maximal matching using quantum annealing. It should be noted that each maximal matching uses two types:
\begin{description}
\item[tracking1.] Similar to SORT, the Intersection over Union (IoU) between the trajectory forecasting by the Kalman filter and the detection location on the current frame is used as the weight. 
\item[tracking2.] The similarity of appearance features between the most recent detected object (object1) in track and the detected 
object (object2) on the current frame, calculated by perceptual hash~\cite{\PCHA{2}}, is used as the weight.
Specifically, let PH1 and PH2 denote the 64-bit Perceptual Hashes that can be obtained from object1 and object2, respectively. When the Hamming distance between PH1 and PH2 is denoted as $d_H ({\rm PH}1, {\rm PH}2)$, the similarity $w_{1,2}$ between object1 and object2 is defined as
\begin{equation}
w_{1,2} := \exp\left( -\frac{d_H({\rm PH}1, {\rm PH}2)}{32} \right).
\end{equation}
\end{description}
\end{minipage}
\end{itemize}
In the proposed method, quantum annealing calculations are configured to perform 100 trials per execution, each with an annealing time 
of  $10\ \mu s$.

\subsubsection{Evaluation Criteria}

Here, each of the following quantitative evaluation metrics for the accuracy of multi-object tracking is discussed: IDSW, 
MOTA~\cite{\MOTAA{2}}, IDF1~\cite{\IDFA{2}}, and the absolute percentage error of vehicle count.

\paragraph{IDSW}

IDSW (ID Switches) is the toal number of times that the identifier (ID) of each object is erroneously switched during tracking.

\paragraph{MOTA}

MOTA (Multiple Object Tracking Accuracy)~\cite{\MOTAA{2}} is a metric that evaluates the frequency of three typical error sources in 
MOT: misses, false positives, and mismatch errors, also known as non-detection, mis-detection, and swapping of tracking IDs. The value range is from $0$ to $1$; the closer to $1$, the greater the tracking performance.
\begin{equation}
{\rm MOTA} := 1- \frac{{\rm FP} + {\rm FN} + {\rm IDSW}}{{\rm GT}}
\end{equation}
where
\begin{itemize}
\item ${\rm FP}$ (False Positives) is the total number of objects that are erroneously detected.
\item ${\rm FN}$ (False Negatives) is the total number of objects that are undetected.
\item ${\rm GT}$ (Ground Truths) is the total number of objects that are annotated.
\end{itemize}

\paragraph{IDF1}

IDF1~\cite{\IDFA{2}} is an evaluation metric for object tracking that measures not only the consistency of tracking, but also the consistency of identification. It is defined by the following formula:

\begin{equation}
{\rm IDF1} = \frac{2\  {\rm IDTP}}{2\  {\rm IDTP} + {\rm IDFP} + {\rm IDFN}}
\end{equation}
where
\begin{itemize}
\item ${\rm IDTP}$ (ID True Positives) is the number of objects that are correctly tracked. A tracking result is considered correct if it matches well with the ground truth data of the object's trajectory.
\item ${\rm IDFP}$ (ID False Positives) is the number of objects that are wrongly tracked.
\item ${\rm IDFN}$ (ID False Negatives) is the number of objects that are missed by the tracker.
\end{itemize}

Similar to the F1-score in object detection, IDF1 is defined as the harmonic mean of ID-Precision=IDTP/(IDTP+IDFP) and ID-Recall=IDTP/(IDTP+IDFN).
IDF1 ranges from 0 to 1, and a higher value indicates a higher accuracy of tracking.

\paragraph{Absolute Percentage Error (APE) of Vehicle Count}

From the perspective of traffic flow measurement, the absolute percentage error of vehicle count is introduced as an intuitively understandable metric:
\begin{equation}
{\rm Absolute\ Percentage\ Error\ of\ Vehicle\ Count}  = \frac{ |{\rm NT} - {\rm CN}| }{ {\rm CN} }
\end{equation}
where
\begin{itemize}
\item ${\rm NT}$ is the number of tracked vehicles
\item ${\rm CN}$ is the correct number of vehicles
\end{itemize}
This metric quantifies the error magnitude as a percentage relative to the correct number of vehicles, indicating that a value closer to 0.0 signifies superior MOT performance.

\subsection{Evaluation Results}

This subsection presents the quantitative evaluation results of the baseline and proposed methods for MOT in Table~\ref{table:evaluation_QAtracking}. It also shows the quantitative evaluation results of the object detection in the pre-processing stage. The ground truth data are obtained from the annotation results of vehicle detection and tracking (vehicle bounding boxes and IDs).

The proposed methods (QA$k$ ($k=1,2,12$)) generally outperform the conventional method (DeepSORT) in all metrics. For the 
movie MVI 39271, the tracking method QA2 based on image features has worse IDSW and APE values than QA1, due to the presence of many taxis with similar appearance. However, the integrated method QA12 can mitigate the errors of QA2 by combining QA1 and QA2.

For the movie MVI 39401, the tracking accuracy of QA1 and QA2 is slightly degraded by the occlusion caused by poles and large vehicles, and the interference of surrounding vehicles with similar appearance. However, the integrated method QA12 can complement the errors of QA1 and QA2 by fusing them. These results demonstrate the effectiveness of the proposed integration method.

\begin{table}[t]
\caption{
Evaluation results of object detection (YOLOv5) and MOT (DeepSORT, Quantum Annealing ) using the UA-DETRAC dataset. It should be noted that the same object detection results were used for all multi-object tracking. Furthermore, QA is an abbreviation for Quantum Annealing, and QA$k$ ($k=1,2$) represents the results obtained by tracking$k$ as explained in 
\ref{subsubsec:setting} Experimental Settings.
QA-12 denotes the integrated results of tracking1 and tracking2.
Values in \textbf{bold} represent the best numerical results among the compared tracking methods.}\label{table:evaluation_QAtracking}
\centering
\begin{tabular*}{13cm}{@{\extracolsep{\fill}}cc|cccc}
\hline
\multicolumn{2}{c|}{\rule{0pt}{2.5ex}\textbf{movie}}     & \multicolumn{4}{c}{\textbf{MVI 39271}}         \\ \hline \hline
%\multirow{3}{*}{\vspace{-1zw}\hspace{2zw}\scalebox{1.2}{detection}}
\multirow{3}{*}{\rule{0pt}{3ex}\large{detection}} & {\rule{0pt}{2.5ex}Precision} & \multicolumn{4}{c}{0.9997}     \\ \cmidrule{2-6} 
                           & Recall    & \multicolumn{4}{c}{0.9143}       \\ \cmidrule{2-6} 
                           & F1-score   & \multicolumn{4}{c}{0.9551}    \\ \hline \cmidrule{1-6}  
%\multirow{7}{*}{\vspace{-4zw}\hspace{2zw}\scalebox{1.2}{tracking}}  
\multirow{7}{*}{\rule{0pt}{7ex}\large{tracking}} & \textbf{method}    & \textbf{DeepSORT} & \textbf{QA1}     & \textbf{QA2}
        & \textbf{QA12}           \\ \cmidrule{2-6} 
                           & MOTA      & 0.9031   & \textbf{0.9050} & 0.9044          & \textbf{0.9050}  \\ \cmidrule{2-6} 
                           & IDF1      & 0.9446   & 0.9513          & \textbf{0.9520} & 0.9513  \\ \cmidrule{2-6} 
                           & IDSW      & 9        & \textbf{3}      & 5               & \textbf{3}      \\ \cmidrule{2-6} 
                           & APE       & 0.1702   & \textbf{0.0000} & 0.0426          & \textbf{0.0000}  \\ \cmidrule{2-6} 
                           & NT      & 55       & \textbf{47}     & 49              & \textbf{47}        \\ \cmidrule{2-6} 
                           & (CN)      & \multicolumn{4}{c}{(47)}          \\ \hline
\end{tabular*}
%\vspace{3zw}
\begin{tabular*}{13cm}{@{\extracolsep{\fill}}cc|cccc}
\hline
\multicolumn{2}{c|}{\rule{0pt}{2.5ex}\rule{0pt}{2.5ex}\textbf{movie}}           & \multicolumn{4}{c}{\textbf{MVI 39401}}     \\ \hline \hline
%\multirow{3}{*}{\vspace{-1zw}\hspace{2zw}\scalebox{1.2}{detection}} 
\multirow{3}{*}{\rule{0pt}{3ex}\large{detection}}  &  {\rule{0pt}{2.5ex}Precision}    & \multicolumn{4}{c}{0.9967}   \\ \cmidrule{2-6} 
                           & Recall      & \multicolumn{4}{c}{0.9528}      \\ \cmidrule{2-6} 
                           & F1-score    & \multicolumn{4}{c}{0.9742}     \\ \hline \cmidrule{1-6}  
%\multirow{7}{*}{\vspace{-4zw}\hspace{2zw}\scalebox{1.2}{tracking}}  
\multirow{7}{*}{\rule{0pt}{7ex}\large{tracking}} & \textbf{method}    & \textbf{DeepSORT} & \textbf{QA1}        & \textbf{QA2}         & \textbf{QA12}            \\ \cmidrule{2-6} 
                           & MOTA       & 0.9424   & 0.9433 & \textbf{0.9436} & 0.9405 \\ \cmidrule{2-6} 
                           & IDF1      & 0.9285   & 0.9563          & \textbf{0.9710} & 0.9646          \\ \cmidrule{2-6} 
                           & IDSW       & 20       & 16     & 15              & \textbf{14}     \\ \cmidrule{2-6} 
                           & APE      & 0.0976   & 0.0244 & 0.0244          & \textbf{0.0000} \\ \cmidrule{2-6} 
                           & NT      & 90       & 84     & 84              & \textbf{82}     \\ \cmidrule{2-6} 
                           & (CN)        & \multicolumn{4}{c}{(82)}         \\ \hline
\end{tabular*}
\end{table}

\clearpage

\section{Proposed Method 2: Efficiency Improvement of MOT by Reverse Annealing}\label{sec:RAMOT}

In MOT, tracking processing is typically performed for each detection frame. However, given that the position of an object changes gradually, the tracking result for the current frame is often similar to that of the previous frame, depending on the situation. 
In this section, by utilizing such properties, a method is proposed to further enhance the efficiency of MOT processing through reverse annealing.

\subsection{Reverse Annealing (RA)}

RA is a technique that employs the inverse process of quantum annealing to search for a more optimal solution when a candidate for the optimal solution is known through other methods~\cite{\RAA{2}}. It is utilized for local solution search. 
In this context, quantum annealing described in subsection~\ref{subsec:forward-algorithm} is called forward annealing (FA) to distinguish it from 
RA.

\begin{figure}[b]
 \centering
 \includegraphics[bb=0 0 389 288, width=9.0cm]{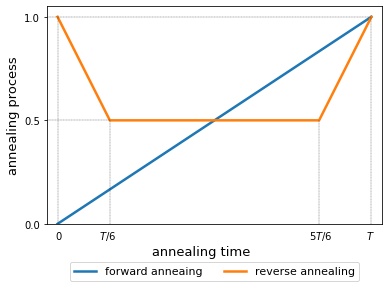}
 \caption{
 An example of an annealing schedule. Here, both forward annealing and reverse annealing are set to the same annealing time $T$. 
 }\label{fig:annesche}
\end{figure}

In the time evolution of the Hamiltonian (Equation \eqref{engeq:hamiltonian_forward}), FA starts from a strong quantum superposition state with a transverse field to select and evaluate all combinations of solutions with equal probability, and gradually reduces the influence of the transverse field. In contrast, RA begins from a state unaffected by the transverse field, with the trial optimal solution as the initial value. It then increases the influence of the transverse field to include the effect of quantum fluctuations. After a temporary pause, it reduces the influence of the transverse field to efficiently search for a more refined solution~\cite{\RAA{2}}. This pause is used in open systems like D-Wave machines that are subject to thermal effects, to efficiently search for solutions using thermal relaxation. Examples of annealing schedules for forward annealing and reverse annealing are shown in Figure~\ref{fig:annesche}.

\subsection{Enhanced MOT through RA}\label{subsec:MOTRA}

RA is utilized to seek a more optimal solution. There have been instances where RA  has expedited the search for solutions with high accuracy using shorter annealing schedules than those in QA (FA)~\cite{\RAC{2},{\RAD{2}}}. 
This paper extends this approach and presents a method that employs reverse annealing to enhance the efficiency of MOT.

\begin{figure}[t]
  \begin{minipage}[b]{0.45\hsize}
    \centering
    \includegraphics[bb=0 0 650 396, width=6.5cm]{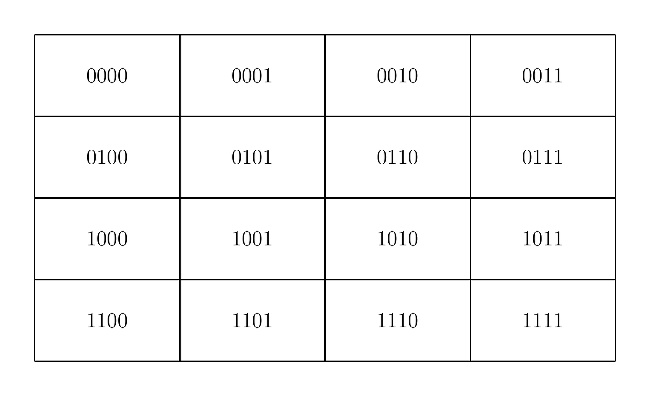}
    \subcaption{location code table}%\label{label1}
  \end{minipage}
  \begin{minipage}[b]{0.45\hsize}
    \centering
    \includegraphics[bb=0 0 650 319, width=6.5cm]{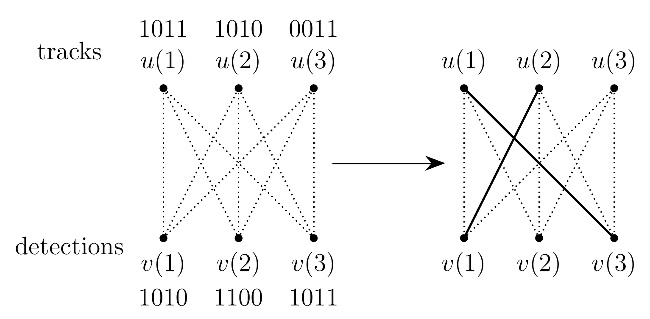}
    \subcaption{construction of initial value}%\label{label2}
  \end{minipage}
  \caption{(a) Example of location code table (when the image is divided into four parts both vertically and horizontally), and
(b) construction of initial value from local code (The four-digit number, noted adjacent to the nodes of the graph on the left, represents 
the position code.).}\label{fig:localcode}
\end{figure}

To rapidly determine the initial values for MOT between the $(t-1)$-th and $t$-th detection frames, each node $u$ of the tracks up to the $(t-1)$-th detection frame and each node $v$ of the detection in the $t$-th frame are encoded as follows. Initially, 
the frame image is partitioned into regions both vertically and horizontally, with a position code assigned to each region (Figure \ref{fig:localcode} (a)). For each node $u$ of the tracks, a position code $c(u)$, corresponding to the center of the bounding box that represents the predicted position in the $t$-th frame, is assigned. This prediction is based on the detection and tracking results up to the $(t-1)$-th detection frame and is made by the Kalman filter in {\bf tracking1.} method described in Subsection ~\ref{subsec:evalexp}. For each node $v$ of the detection, a position code $c(v)$, corresponding to the center of the bounding box that represents the detection result, is assigned. Only when $c(u)=c(v)$, a subgraph that matches $u$ and $v$ is constructed and used as the initial value (Figure \ref{fig:localcode} (b)). When formulating the 
MOT problem between the $(t-1)$-th and $t$-th detection frames in the same manner as Equations \eqref{engeq:QUBO} and \eqref{engeq:integQUBO}, the initial value
$\widehat{\bm x}=\{ x_{u,v} \mid (u,v)\in E \}$ can be expressed as follows:
\begin{equation}
x_{u,v} = 
\begin{cases}
1 & {\rm as} \ \ c(u)=c(v)  \\
0 & {\rm otherwise}
\end{cases}
\end{equation}
Starting from this initial value, a more refined solution is sought by RA.

\begin{figure}[t]
  \begin{minipage}[b]{0.33\hsize}
    \centering
    \includegraphics[bb=0 0 389 292, width=5.5cm]{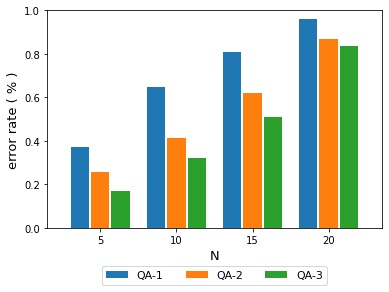}
    \subcaption{QA}%\label{label1}
  \end{minipage}
  \begin{minipage}[b]{0.33\hsize}
    \centering
    \includegraphics[bb=0 0 389 292, width=5.5cm]{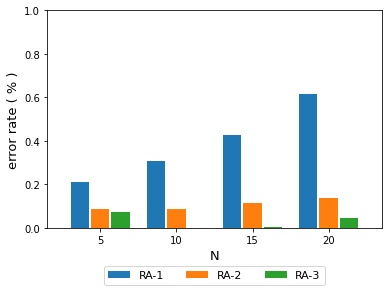}
    \subcaption{RA-(1)}%\label{label2}
  \end{minipage}
  \begin{minipage}[b]{0.33\hsize}
    \centering
    \includegraphics[bb=0 0 389 292, width=5.5cm]{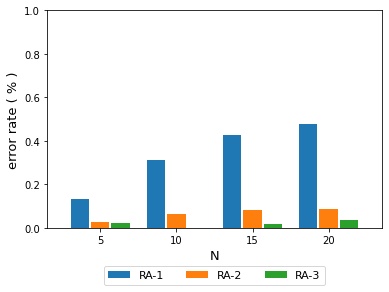}
    \subcaption{RA-(2)}%\label{label2}
  \end{minipage}
  \caption{Comparison of error rates by QA (Quantum Annealing) and RA (Reverse Annealing).
The multiplicity $k$ of the integration of matching solutions is denoted as QA-$k$ and RA-$k$. The annealing time per trial in 
(b) RA-(1) and (c) RA-(2) was 10 $\mu$s and 3 $\mu$s, respectively.}\label{fig:ErrorRates_QA_RA}
\end{figure}

To verify the effect of maximal matching by reverse annealing, the same artificially generated $N \times N$ bipartite graph as used in Subsection~\ref{subsec:integ_matching} is utilized. 
The results of the quantitative evaluation by error rate are shown in Figure ~\ref{fig:ErrorRates_QA_RA}. 
Trial computations were conducted 250 times per graph. 
The initial values for reverse annealing were obtained by randomly flipping the bits of the optimal solution for maximal matching with a probability of 5\%. This was premised on the assumption that the initial value predictions, based on the position prediction by the tracker and the object detection results, would be incorrect at a rate of less than 5\% per quantum bit.

From these results, it can be anticipated that if the initial value can be predicted almost accurately, the accuracy of matching by reverse annealing can be expected to be higher than that of forward annealing. By integrating the results of multiple matchings, a further improved accuracy can be expected.

\begin{figure}[t]
  \begin{minipage}[b]{0.33\hsize}
    \centering
    \includegraphics[bb=0 0 408 288, width=5.5cm]{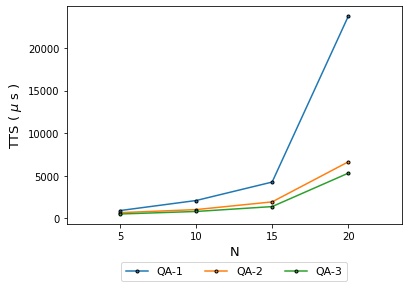}
    \subcaption{QA}%\label{label1}
  \end{minipage}
  \begin{minipage}[b]{0.33\hsize}
    \centering
    \includegraphics[bb=0 0 396 292, width=5.5cm]{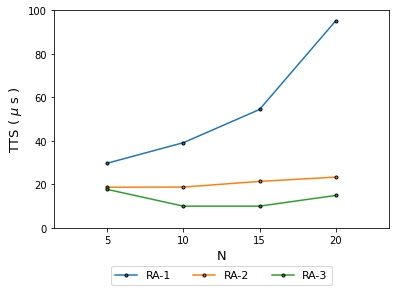}
    \subcaption{RA-(a)}%\label{label2}
  \end{minipage}
  \begin{minipage}[b]{0.33\hsize}
    \centering
    \includegraphics[bb=0 0 396 292, width=5.5cm]{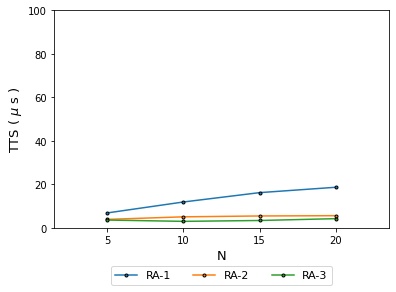}
    \subcaption{RA-(b)}%\label{label2}
  \end{minipage}
  \caption{Comparison of TTS by QA and RA}\label{fig:TTS_QA_RA}
\end{figure}

Also, similar to Figure ~\ref{fig:TTS_QA_RA}, from the results evaluated by TTS, it can be inferred that the solution accuracy per unit time of reverse annealing is better than that of forward annealing, and the correct answer can be obtained with fewer trials of annealing computation.

\subsection{Evaluation Experiments}

This Subsection describes the results of evaluation experiments carried out to demonstrate the effectiveness of MOT by RA.

\subsubsection{Experimental Data and Settings} 

The same experimental data and settings as described in subsection~\ref{subsec:evalexp} was used.
The annealing time for RA was set to $3\ \mu$s, and the number of repeated trials per tracking process was limited to one.

\subsubsection{Evaluation Results}

\begin{table}[t]
\caption{The accuracy when performing MOT by RA, using the same tracking results as in Table \ref{table:evaluation_QAtracking}.
RA$k$ ($k=1,2$) represents the results obtained by tracking$k$ as explained in 
\ref{subsubsec:setting} Experimental Settings.
RA12 denotes the integrated results of {\bf tracking1} and {\bf tracking2}.}\label{tbl:Tracking_RA}
\centering
\begin{tabular*}{13cm}{@{\extracolsep{\fill}}cc|cccccc}
\hline
\multicolumn{2}{c|}{\rule{0pt}{2.5ex}\textbf{movie}}       & \multicolumn{6}{c}{\textbf{MVI 39271}}                                                                            \\ \hline \hline
%\multirow{7}{*}{\vspace{-4zw}\hspace{2zw}\scalebox{1.2}{tracking}}
\multirow{7}{*}{\rule{0pt}{7ex}\large{tracking}} & %\textbf{method}
{\rule{0pt}{2.5ex}\textbf{method}}  & \textbf{QA1}             & \textbf{QA2}             & \textbf{QA12}            & \textbf{RA1}             & \textbf{RA2}             & \textbf{RA12}            \\ \cmidrule{2-8}
                          & MOTA   & \textbf{0.9050} & 0.9044          & \textbf{0.9050} & 0.9047          & 0.9040          & 0.9047          \\ \cmidrule{2-8} 
                          & IDF1   & 0.9513          & \textbf{0.9520} & 0.9513          & 0.9513          & \textbf{0.9520} & 0.9513          \\ \cmidrule{2-8} 
                          & IDSW   & \textbf{3}      & 5               & \textbf{3}      & \textbf{3}      & 5               & \textbf{3}      \\ \cmidrule{2-8} 
                          & APE    & \textbf{0.0000} & 0.0426 & \textbf{0.0000} & \textbf{0.0000} & 0.0426          & \textbf{0.0000} \\ \cmidrule{2-8} 
                          & NT     & \textbf{47}     & 49              & \textbf{47}     & \textbf{47}     & 49              & \textbf{47}     \\ \cmidrule{2-8} 
                          & (CN)   & \multicolumn{6}{c}{(47)}                                                                                  \\ \hline
\end{tabular*}
%\vspace{3zw}
\begin{tabular*}{13cm}{@{\extracolsep{\fill}}cc|cccccc}
\hline
\multicolumn{2}{c|}{\rule{0pt}{2.5ex}\textbf{movie}}       & \multicolumn{6}{c}{\textbf{MVI 39401}}                                                                            \\ \hline \hline
%\multirow{7}{*}{\vspace{-4zw}\hspace{2zw}\scalebox{1.2}{tracking}}
\multirow{7}{*}{\rule{0pt}{7ex}\large{tracking}} & %\textbf{method} 
{\rule{0pt}{2.5ex}\textbf{method}}  & \textbf{QA1}             & \textbf{QA2}             & \textbf{QA12}            & \textbf{RA1}             & \textbf{RA2}             & \textbf{RA12}            \\ \cmidrule{2-8}
                          & MOTA   &  0.9433 & \textbf{0.9436} & 0.9405 & 0.9402          & 0.9402          & 0.9409          \\ \cmidrule{2-8} 
                          & IDF1   & 0.9563      & 0.9710 & 0.9646          & 0.9563          & \textbf{0.9712} & 0.9366          \\ \cmidrule{2-8} 
                          & IDSW   & 16   & 15              & 14   & 16    & 14              & \textbf{12}     \\ \cmidrule{2-8} 
                          & APE    & 0.0244 & 0.0244 & \textbf{0.0000} & 0.0244 & 0.0244      & \textbf{0.0000} \\ \cmidrule{2-8} 
                          & NT     & 84     & 84              & \textbf{82}     & 84     & 84              & \textbf{82}     \\ \cmidrule{2-8} 
                          & (CN)   & \multicolumn{6}{c}{(82)}                                                                                  \\ \hline
\end{tabular*}
\end{table}

The same object tracking results and evaluation criteria as in Subsection~\ref{subsec:evalexp} are used and presented in 
Table~\ref{tbl:Tracking_RA}. Although the results are comparable to those of QA, it can be observed from the results using the video MVI 39401 that the method RA12 has the least IDSW, among others. This indicates that good results can be obtained with a short annealing time 
(3 $\mu$s per tracking process).

\begin{figure}[t]
 \centering
\includegraphics[bb=0 0 584 266, width=13cm]{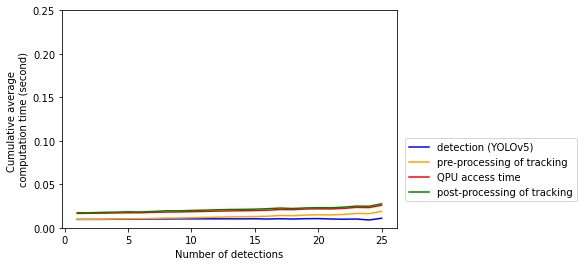}
 \caption{The cumulative processing time for detection by YOLOv5 and tracking by RA111 (triple matching). The right axis represents the number of detected vehicles, and the vertical axis represents the total processing time for detection and tracking (unit: seconds).
}\label{fig:cumulative_sum_YOLO_RA}
\end{figure}

In the method of RA12, the error rate per quantum bit of the initial value predicted using the location code described in 
Subsection~\ref{subsec:MOTRA} was 2.76\%. 
In the method of RA111 (triple matching), the QPU access time per tracking process had an average value (AV) of 6.66 ms and a standard deviation (SD) of 1.33 ms. The pre-processing time for tracking, which includes initial value prediction, had an AV of 2.60 ms and a SD of 2.29 ms. The post-processing time for tracking, which includes integration processing, had an AV of 1.27 ms and a SD of 0.38 ms.
The cumulative sum of these processing times and the tracking processing times is depicted in Figure~\ref{fig:cumulative_sum_YOLO_RA}.

\section{Conclusions}\label{sec:CONC}

In this paper, a method that leverages quantum annealing to enhance both the speed and accuracy of MOT by ensembling the 
MOT process was proposed and demonstrated. 
A method that further improves the efficiency of MOT by using RA was also proposed and demonstrated. Although there are still challenges in implementing QA on edge computing platforms and ensuring its cost-effectiveness, it is expected that proposed
methods will be useful for real-time MOT applications, such as traffic flow measurement and control, collision prediction for autonomous driving, and quality control of mass-produced products. It is anticipated that proposed
methods will contribute to the advancement of next-generation transportation and automation fields.

\section*{Acknowledgments}

In the execution of this research, substantial assistance was received from colleagues at 
NEC Solution Innovators, Ltd. The author express the profound gratitude.

\clearpage

\bibliographystyle{unsrtnat}
\bibliography{qmot_paper}

\end{document}